\documentclass[sigconf]{acmart}
\AtBeginDocument{%
  \providecommand\BibTeX{{%
    \normalfont B\kern-0.5em{\scshape i\kern-0.25em b}\kern-0.8em\TeX}}}

\setcopyright{acmcopyright}
\copyrightyear{2018}
\acmYear{2018}
\acmDOI{XXXXXXX.XXXXXXX}

\acmConference[Conference acronym 'XX]{Make sure to enter the correct
  conference title from your rights confirmation emai}{June 03--05,
  2018}{Woodstock, NY}
\acmPrice{15.00}
\acmISBN{978-1-4503-XXXX-X/18/06}

\begin{document}

\title{TEDL: A Two-stage Evidential Deep Learning Method for Classification Uncertainty Quantification}

% \affiliation{%

\author{Xue Li}
\affiliation{%
  \institution{Microsoft}
  \country{USA}
  }
\email{xeli@microsoft.com}

\author{Wei Shen}
\affiliation{%
  \institution{Microsoft}
  \country{USA}
}
\email{sashen@microsoft.com}

\author{Denis Charles}
\affiliation{%
  \institution{Microsoft}
  \country{USA}
}
\email{cdx@microsoft.com}


\begin{abstract}
  In this paper, we propose TEDL, a two-stage learning approach to quantify uncertainty for deep learning models in classification tasks, inspired by our findings in experimenting with Evidential Deep Learning (EDL) method, a recently proposed uncertainty quantification approach based on the Dempster-Shafer theory. 
  More specifically, we observe that EDL tends to yield inferior AUC compared with models learnt by cross-entropy loss and is highly sensitive in training. Such sensitivity is likely to cause unreliable uncertainty estimation, making it risky for practical applications. To mitigate both limitations, we propose a simple yet effective two-stage learning approach based on our analysis on the likely reasons causing such sensitivity, with the first stage learning from cross-entropy loss, followed by a second stage learning from EDL loss. We also re-formulate the EDL loss by replacing \textit{ReLU} with \textit{ELU} to avoid the \textit{Dying ReLU} issue. Extensive experiments are carried out on varied sized training corpus collected from a large-scale commercial search engine, demonstrating that the proposed two-stage learning framework can increase \textit{AUC} significantly and greatly improve training robustness.
\end{abstract}



\maketitle

%--------------------------------------------------------------------------------------------------------------------

\section{Introduction}
\label{sec_intro}

% \begin{figure}[!t]
% \begin{center}
%   \includegraphics[width = 1.0\linewidth]{figures/overview.png}
% \end{center}
%   \caption{A schematic illustration of the proposed TEDL method. 
%   (a) The original EDL method transforms the model outputs to strictly positive values using $\textit{ReLU}$ activation and learns to quantify uncertainty via the EDL loss in Equation (\ref{eqn.edl_final_loss}), which yields inferior AUC and is sensitive to training. 
%   (b) The proposed TEDL method employs a two-stage learning strategy to decompose the original problem into two easier sub-problems and tackle one at a time: the first stage learns to make good pointwise estimations via cross-entropy loss; and then the second stage will learn to quantify uncertainty using the pointwise estimation as anchor points, with \textit{ReLU} replaced by \textit{ELU} to avoid the \textit{Dying ReLU} issue. Compared with EDL, TEDL achieves significantly higher AUC and is much more robust.}
%   \label{fig.overview}
% \end{figure}

\begin{figure*}[!t]
\begin{center}
   \includegraphics[width = 0.8\linewidth]{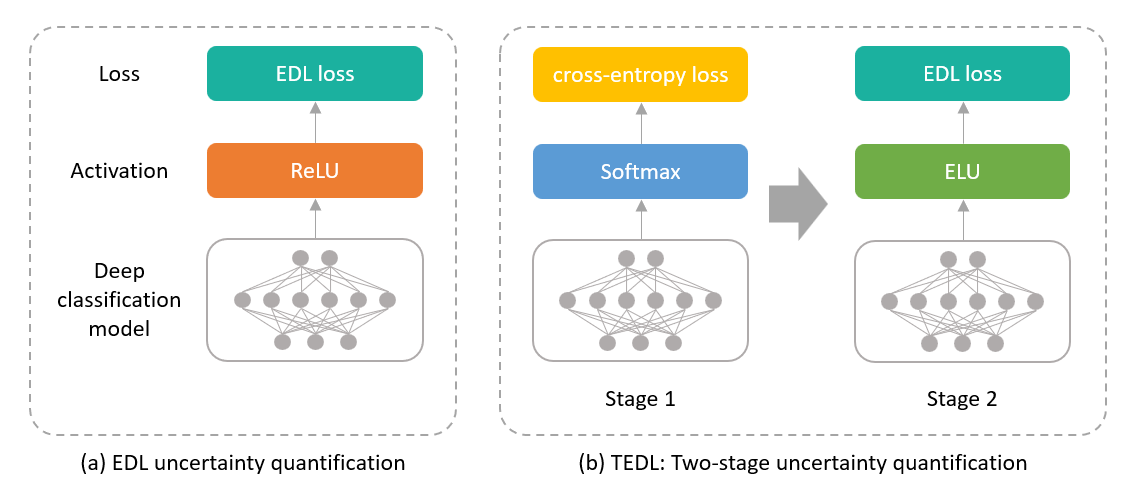}
\end{center}
   \caption{A schematic illustration of the proposed TEDL method. 
   (a) The original EDL method transforms the model outputs to strictly positive values using $\textit{ReLU}$ activation and learns to quantify uncertainty via the EDL loss in Equation (\ref{eqn.edl_final_loss}), which yields inferior AUC and is sensitive to training. 
   (b) The proposed TEDL method employs a two-stage learning strategy to decompose the original problem into two easier sub-problems and tackle one at a time: the first stage learns to make good pointwise estimations via cross-entropy loss; and then the second stage will learn to quantify uncertainty using the pointwise estimation as anchor points, with \textit{ReLU} replaced by \textit{ELU} to avoid the \textit{Dying ReLU} issue. 
   %Compared with EDL, TEDL achieves significantly higher AUC and is much more robust.
   }
   \label{fig.overview}
\end{figure*}

Uncertainty quantification of deep learning models has been a hot topic in the community ever since the rise of deep learning, and the demand for effective uncertainty quantification methods is becoming increasingly urgent in the recent decade as deep learning continue to reshape many industries. Search recommendation, as perhaps the most radically reshaped industry, often relies on many different deep learning models to give accurate recommendations, which makes uncertainty quantification especially important since unreliable predictions could accumulate in the system and finally lead to inaccurate or even embarrassing recommendation results.

To make machine learning models aware of their own prediction confidence, many uncertainty quantification approaches have been proposed \cite{gawlikowski2021survey}, including single deterministic methods, Bayesian methods and ensemble methods, etc., among which the single deterministic methods could be further grouped into internal or external methods depending on whether additional components are required for uncertainty estimation. We present a brief review on this topic in Section \ref{sec_related}. In this paper, we are particularly interested in single deterministic methods, especially internal approaches, since such methods typically need only a single forward pass on a deterministic network to estimate uncertainty, and hence does not require stochastic DNN or ensemble models, making both training and inference more efficient. 

More specifically, instead of considering the model outputs as a pointwise maximum-a-posteriori (MAP) estimation, internal single deterministic methods usually interpret model outputs as parameters of a prior distribution over all the possible predictions, and then give prediction by taking the expected value over the prior distribution. For classification tasks, Dirichlet distribution is often chosen as prior since it is the conjugate prior of the categorical distribution. Meanwhile, statistical distance metrics such as Kullback-Leibler (KL) divergence are often included in their loss functions due to the need to optimize on parameters of distributions \cite{sensoy2018evidential, malinin2018predictive}. 

However, the efficiency of such methods comes with a cost. As mentioned in \cite{gawlikowski2021survey}, they are typically more sensitive towards training settings such as initialization, hyper-parameters, training data, etc., which is what we observed when apply EDL \cite{sensoy2018evidential}, a recently proposed single deterministic method, to practical scenarios.

% More specifically, the authors of EDL firstly point out that the \textit{softmax} outputs should be interpreted as the point estimate of the categorical distribution, and starting from this point, they place a Dirichlet prior over all the possible \textit{softmax} outputs and derive a loss function called EDL loss following the Dempster-Shafer theory \cite{dempster1968generalization}. In addition, they also introduce a regularization term to punish features providing evidence for multiple categories 
% by minimizing KL divergence between the Dirichlet prior and the uniform distribution, along with its associated coefficient $\lambda_t$, which is heuristically set to increase gradually with epoch $t$ to avoid premature convergence to the uniform distribution.

\begin{figure*}[!t]
\begin{center}
   \includegraphics[width = 1.0\linewidth]{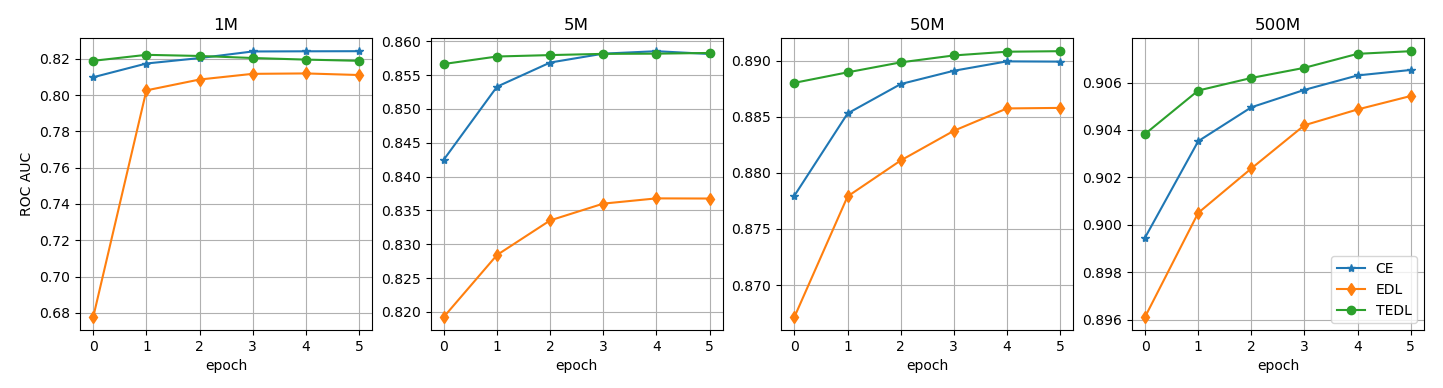}
\end{center}
   \caption{AUC comparison between cross-entropy loss, EDL loss and TEDL loss, evaluated on the same validation data. All the three methods are learnt on training corpus with 1M, 5M and 50M samples, respectively. In all the training settings, the EDL method achieves inferior AUC compared to cross-entropy loss, while the proposed TEDL method yields comparable AUC than cross-entropy, outperforming EDL significantly.}
   \label{fig.exp_auc}
\end{figure*}

To be more specific, in our experiments we identify several issues in the EDL method. Firstly, as shown in Figure \ref{fig.exp_auc}, when applied to binary classification tasks, the \textit{ROC AUC} achieved by the EDL method is significantly lower than that obtained by cross-entropy loss, and such gap cannot be bridged by simply adding more training samples. Secondly, EDL tends to be sensitive to initialization and some hyper-parameters, where improper settings may lead to significantly degraded \textit{AUC} and unreliable uncertainty estimation. 

To see this more clearly, in Figure \ref{fig.sensitivity_auc} (the orange curve) we summarize the per epoch \textit{ROC AUC} obtained in EDL training with different $\lambda$, a hyper-parameter controlling how close the Dirichlet prior is to a uniform distribution. As we can see, the \textit{AUC} of EDL suffers in the beginning under all the four settings, and in some cases (for example when $\lambda=0.5$) there is no signs of improvement at all. In cases where \textit{AUC} does improve, its final \textit{AUC} is still significantly lower than that from the proposed method (the green curve). On the other hand, consider evaluating \textit{AUC} on validation samples with uncertainty lower than a certain threshold: If the learnt uncertainty is of high quality, smaller thresholds should indicate higher confidence, and hence should be associated with higher \textit{AUC}. However, this is not always the case for EDL, as shown in the first row of Figure \ref{fig.sensitivity_auc_u}. Besides, we also observe that when a large $\lambda$ is used (for example $\lambda$=0.75), there would be a higher risk of running into the \textit{Dying ReLU} problem where all outputs are zero, leading to an \textit{AUC} that similar to a random guess. All these issues make it risky to apply methods like EDL into real-world applications.

To fix these issues, we firstly present an analysis in this paper on the likely reasons causing the above issues in Section \ref{sec_approach}, and based on our analysis, we further propose \textbf{TEDL}, short for \textbf{T}wo-stage \textbf{E}vidential \textbf{D}eep \textbf{L}earning, as a simple but effective training framework to mitigate all the aforementioned issues in a single shot. 
As we will see in Section \ref{sec_approach}, the basic idea of TEDL is to transform the difficult uncertainty quantification problem into two sub-problems that are much easier to tackle, i.e., 1) finding a reasonably good pointwise estimation of the categorical distribution, and 2) leveraging this pointwise estimation as an anchor point for estimating the Dirichlet prior of categorical distribution, based on which we can quantify uncertainty.

The overall training framework of TEDL is illustrated in Figure \ref{fig.overview}, where two stages are needed: in the first stage, we train our classification model with cross-entropy loss, in order to obtain a model that is able to output reasonable pointwise estimations of the categorical distribution. And then in the second stage, we initialize the model from the weights obtained in the previous stage, and go through the same training corpus by learning with the reformulated EDL loss where \textit{ReLU} is replaced by \textit{ELU}. As shown in Section \ref{sec_exp}, compared with the EDL baseline, TEDL can achieve higher \textit{AUC} across all evaluation settings and effectively avoid the risk of running into \textit{Dying ReLU} problem. More importantly, TEDL also shows significantly improved robustness towards training settings, making it more reliable for practical applications. 

It is also worth to mention that we name our proposed method following EDL mainly due to the convenience of experimentation, as it is proposed recently and is easy to implement with code open-sourced by the authors. However, our analysis in Section \ref{sec_approach} also applies to other single deterministic uncertainty quantification methods suffering from similar issues, and hence the two-stage learning framework we propose in this paper could be readily extended to those methods as well. 

% Our contributions in this paper could be summarized as follows:
% \begin{itemize}
%     \item we present an analysis on the reasons why single deterministic uncertainty quantification methods tend to be sensitive towards initialization and hyper-parameters, and based on our analysis, we propose TEDL, a two-stage learning method to quantify classification uncertainty for deep learning models, based on the recently proposed EDL method. The two-stage learning framework we propose in this paper could be readily extend to other uncertainty quantification methods suffering from similar issues. 
%     \item we conduct extensive experiments on varied sized training corpus sampled from real commercial search engine, under different settings of hyper-parameters. Our experiments demonstrates the effectiveness of the proposed two-stage learning framework, shedding light upon future works in uncertainty quantification.
% \end{itemize}

%--------------------------------------------------------------------------------------------------------------------

\section{Related Works}
\label{sec_related}

% %--------------------------------------------------------------------------------------------------------------------

The interest for uncertainty estimation dates back to the days even before the rise of deep learning, entailing a large body of literature on this topic. Based on whether model ensemble is used and whether the model is stochastic, uncertainty quantification methods could be roughly grouped into three categories, including single deterministic methods, Bayesian neural networks and ensemble methods. Please refer to \cite{gawlikowski2021survey} for a comprehensive survey.

\textbf{Single deterministic methods} \cite{nandy2020towards,oala2020interval,mozejko2018inhibited} estimate uncertainty based on one single forward pass within a deterministic network, and could be further split into external approach \cite{lee2020gradients,raghu2019direct} and internal approach \cite{sensoy2018evidential,malinin2018predictive,ramalho2020density} depending on whether additional method is used for deriving uncertainty estimation. Methods in this category typically have lower requirements on computational resources since no stochastic networks nor model ensembles are needed, but suffer from sensitivity to initialization and parameters compared with other categories. The proposed TEDL method in this paper, as well as the original EDL method, both fall into this category.   

\textbf{Bayesian neural networks} cover all kinds of stochastic DNNs, including methods based on variational inference \cite{hinton1993keeping,gal2016dropout,blundell2015weight,barber1998ensemble,graves2011practical,louizos2017bayesian,rezende2015variational}, sampling methods \cite{neal1992bayesian,neal1994improved,neal2012bayesian,welling2011bayesian,nemeth2021stochastic}, and Laplace approximation \cite{salimans2016weight,lee2020estimating,ritter2018scalable}. Methods in this category usually have higher computational complexity in both the training and inference phases due to stochastic sampling.

\textbf{Ensemble methods} \cite{lakshminarayanan2017simple,achrack2020multi,huang2017snapshot,cavalcanti2016combining,guo2018margin,martinez2021ensemble,lindqvist2020general,malinin2019ensemble,valdenegro2019deep,wen2020batchensemble} combine the predictions from several different deterministic networks at inference. Methods in this category typically have higher requirements on both the memory and computational resources at inference phase.  

The proposed method also relates to the concept of \textbf{two-stage learning}, which bears similarity to transfer learning but has some subtle differences. Transfer learning generally refers to the procedure that transfers knowledge obtained from different but related source domains to target domains, usually to reduce training data required on the target domains. \cite{zhuang2020comprehensive} gives a comprehensive survey on transfer learning. In contrast, in two-stage learning \cite{dang2013two, khan2019novel}, although it also consists of two consecutive stages, these two stages are often conducted on the same data. In a typical two-stage learning setting, the second stage should be the final stage that yields the desired output, while the first stage serves as a preparation step. Given such differences, the proposed method should be categorized as two-stage learning.

%--------------------------------------------------------------------------------------------------------------------

\section{Approach}
\label{sec_approach}

% This section will begin with a recap of the original EDL method, to give a more intuitive impression on how uncertainty is estimated by internal single deterministic methods. After this, an analysis on their likely cause of sensitivity will be presented, followed by an explanation on why the proposed two-stage learning framework could help mitigate such issues. Following general conventions, vectors will be denoted in bold face in this paper such as $\boldsymbol{x}$, with its $k$-th element denoted as $x_k$.

%--------------------------------------------------------------------------------------------------------------------

\subsection{A Recap on EDL Uncertainty Quantification}
\label{sec_sec_edl}

The basic idea of EDL method is treating \textit{softmax} output as the pointwise estimation of the categorical distribution, and placing a Dirichlet prior over the distribution of all possible \textit{softmax} outputs. Then, following the Dempster-Shafer theory, assume we have $K$ categories and $\boldsymbol{\alpha}_i=\langle \alpha_{i1}, \dots, \alpha_{iK} \rangle$ is the parameter of a Dirichlet distribution for the classification of sample $i$, the authors propose to replace \textit{softmax} with \textit{ReLU} and represent the Dirichlet parameter as $\boldsymbol{\alpha}_i = f(\boldsymbol{x}_i | \Theta) + 1$ where $\Theta$ represents network parameters and $f(\boldsymbol{x}_i | \Theta)$ is the \textit{ReLU} outputs. The $\alpha_{ij}$ here also represents the subjective opinion collected from sample $i$ and category $j$, and $S_i = \sum_{j=1}^{K} \alpha_{ij}$ is referred to as the Dirichlet strength. Note that $S_i$ is inversely proportional to uncertainty: a larger $S_i$ indicates more evidence are collected for sample $i$, and hence less uncertainty.

Based on the above assumptions, the EDL loss is defined as below:

\begin{eqnarray}\label{eqn.edl_final_loss}
\mathcal{L}(\Theta) = \sum_{i=1}^{N} \mathcal{L}_{i}(\Theta) + \lambda_{t} \sum_{i=1}^{N} KL 
 \left[ 
    D(\boldsymbol{p}_i | \tilde{\boldsymbol{\alpha}}_{i}) \hspace{1mm} || \hspace{1mm} 
    D(\boldsymbol{p}_i | \langle 1,\dots, 1 \rangle)
 \right]
\end{eqnarray}
where $\mathcal{L}_{i}(\Theta)$ is formulated as the expected value of a basic loss. According to the authors of \cite{sensoy2018evidential}, EDL method appears relatively more stable when sum of squares loss is used as the basic loss, as below:
\begin{eqnarray}\label{eqn.edl_loss}
\mathcal{L}_{i}(\Theta) &&= \sum_{j=i}^{K}(y_{ij} - \hat{p}_{ij})^2 + \frac{\hat{p}_{ij}(1-\hat{p}_{ij})}{S_i+1} \nonumber \\
&& = \sum_{j=1}^{K} (y_{ij} - \frac{\alpha_{ij}}{S_{i}})^2 + \frac{\alpha_{ij}(S_i - \alpha_{ij})}{S_{i}^2(S_i+1)}
\end{eqnarray}
where $y_{ij}$ and $\hat{p}_{ij}$ denote the class label and expectation for sample $i$ and class $j$, respectively.

Equation (\ref{eqn.edl_final_loss}) also contains a regularization term minimizing the KL divergence between the estimated Dirichlet distribution $D(\boldsymbol{p}_i | \tilde{\boldsymbol{\alpha}}_{i})$ and the uniform distribution. Its associated coefficient $\lambda_t$ is heuristically set to increase with epoch $t$ (zero-based), i.e., $\lambda_t = \min(1.0, t * \lambda)$ where $\lambda=0.1$. Note that we denote the per-epoch increment as $\lambda$. For brevity, we will treat $\lambda$ rather than $\lambda_t$ as the hyper-parameter henceforth, since $\lambda_t$ is determined only by $\lambda$. 

% Previously in Section \ref{sec_intro}, we have seen how $\lambda$ will impact the learning of EDL method and lead to unreliable uncertainty quantification results. In the next section, we will discuss how to fix the issues identified in Section \ref{sec_intro} via two-stage learning.

%--------------------------------------------------------------------------------------------------------------------

\subsection{A Closer Look into the EDL Method}
\label{sec_sec_analysis}

Equation (\ref{eqn.edl_final_loss}) could be split into two parts: the first part is Equation (\ref{eqn.edl_loss}) which is designed to estimate the Dirichlet prior, and the second part is the regularization term derived from KL divergence. Next, we will take a closer look at these two parts respectively to understand the cause of sensitivity.

As we mentioned previously, unlike cross-entropy loss which is designed to learn the pointwise estimations of the categorical distribution as a MAP estimate, the loss function in Equation (\ref{eqn.edl_loss}) is derived to learn the parameter of a Dirichlet prior distribution over all the possible predictions. Therefore, the pointwise estimation should also be covered by the Dirichlet prior distribution. This perspective highlights the huge gap in terms of how difficult the optimization problems behind these two loss functions are, especially given that obtaining a good MAP estimation is already a hard problem in many applications. This perspective also highlights the importance of a sufficiently large training data, as it would be meaningless to model a distribution without sufficient samples.

In the meanwhile, the KL divergence also makes optimization more complicated since it is not Lipschitz smooth. More precisely, given a function $f$, it is said to be Lipschitz smooth if and only if there exists a finite value $L$ such that

% \begin{eqnarray}\label{eqn.lipschitz}
% \frac{\|\nabla f(a) - \nabla f(b)\|}{\|a-b\|} < L
% \end{eqnarray}

\begin{eqnarray}\label{eqn.lipschitz}
\|\nabla f(a) - \nabla f(b)\| < L \cdot \|a-b\|
\end{eqnarray}

In other words, the gradient of $f$ should exist and be bounded by a finite value $L$. However, the regularization term in Equation (\ref{eqn.edl_final_loss}) does not satisfy this condition since its gradient will go to infinity when $D(\boldsymbol{p}_i | \tilde{\boldsymbol{\alpha}}_{i}) \to 0$, as even though $\tilde{\boldsymbol{\alpha}}_{i}$ is guaranteed to be positive, $\boldsymbol{p}_i$ may still become very close to zero when a certain $\tilde{\boldsymbol{\alpha}}_{i}$ is extremely large, leading to very large gradients and hence unstable training.

In summary, internal single deterministic methods are trying to optimize an inherently difficult problem, with potentially ill-conditioned loss functions due to existence of KL divergence.

%--------------------------------------------------------------------------------------------------------------------

\subsection{The Proposed Two-stage Learning Framework}
\label{sec_sec_method}

Having analyzed the possible reasons causing training sensitivity, a more important question is how could we fix such issues and make training more stable. At first glance, this appears to be infeasible since we can neither bypass distribution modeling nor drop the terms related to KL divergence in loss functions. In this paper, we propose an alternative approach, which can fix both issues with a simple yet effective strategy: decomposing the original problem into two sub-problems and tackling one at a time, leading to a two-stage learning method as illustrated in Figure \ref{fig.overview}. Compared with the original EDL method, the only cost introduced by TEDL is a preparation stage learning from the cross-entropy loss, however as we will see in Section \ref{sec_exp}, such cost is well paid off given the significant \textit{AUC} increase and greatly improved robustness in training. 

So why does such a simple strategy work? On one hand, the first stage in TEDL learns a pointwise estimation of the categorical distribution, which is a much easier problem compared with modeling the entire distribution and entails much fewer training samples. Then in the second stage, since the model is initialized from the weights obtained in stage 1, it amounts to modeling the prior distribution using the pointwise estimation as certain anchor points, which is much easier than modeling the prior from scratch, if we can assume that the pointwise estimation is close to the expected value of the prior. This assumption should be easily hold for most practical applications, otherwise we will not be able to apply internal single deterministic methods at all, since the expected value from the prior distribution is unlikely to derive meaningful predictions in that case. 

On the other hand, by learning from cross-entropy loss, we could effectively avoid assigning extremely small values to $\boldsymbol{p}_i$, given that \textit{softmax} involves exponential operations and there is no point in pushing model outputs before \textit{softmax} to extremely large values. That means, when \textit{softmax} is replaced by \textit{ELU} later in stage 2, it is unlikely for us to see extremely large $\tilde{\boldsymbol{\alpha}}_{i}$ values.

\begin{figure*}[!h]
\begin{center}
   \includegraphics[width = 1.0\linewidth]{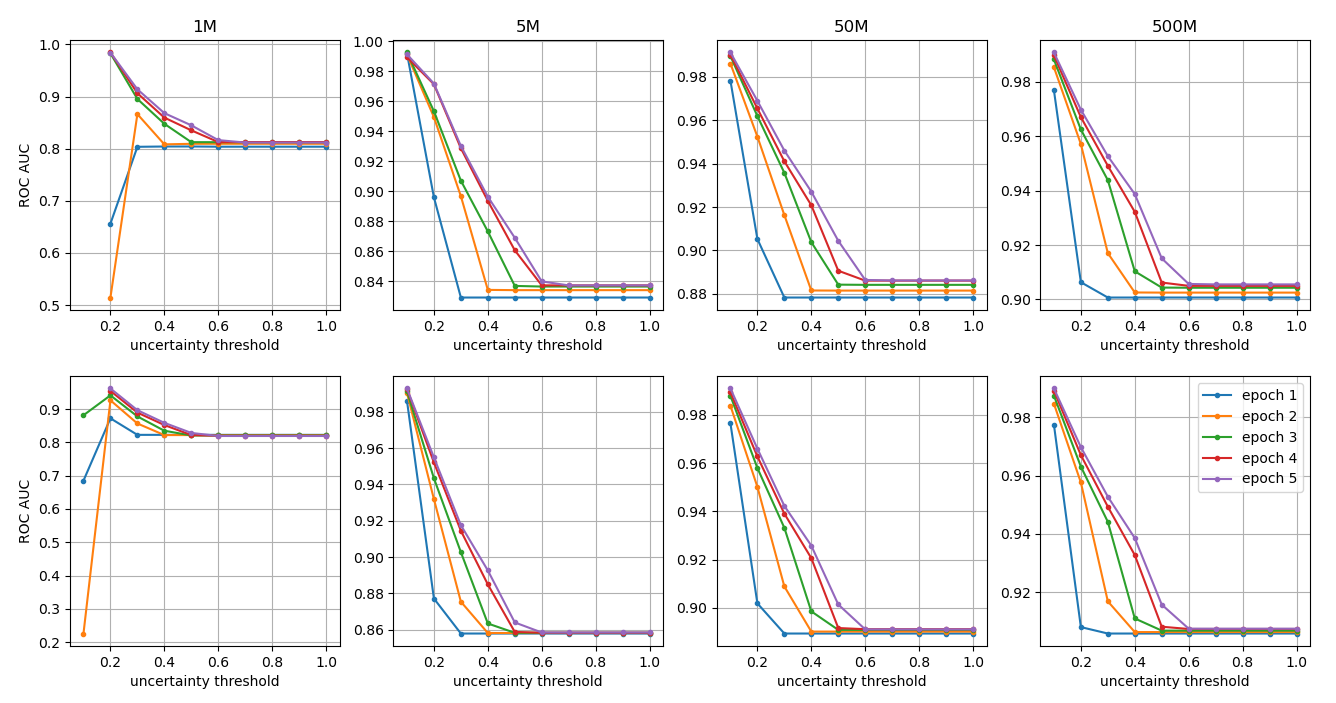}
\end{center}
   \caption{\textit{ROC AUC} vs. uncertainty thresholds with 1M, 5M and 50M training corpus, respectively, and $\lambda = 0.1$. The first row is for EDL, while the second row is for TEDL. This figure shows that under a relatively small $\lambda$, the quality of uncertainty learnt by both EDL and TEDL improves as training proceeds.}
   \label{fig.exp_auc_u}
\end{figure*}

\begin{figure*}[!h]
\begin{center}
   \includegraphics[width = 1.0\linewidth]{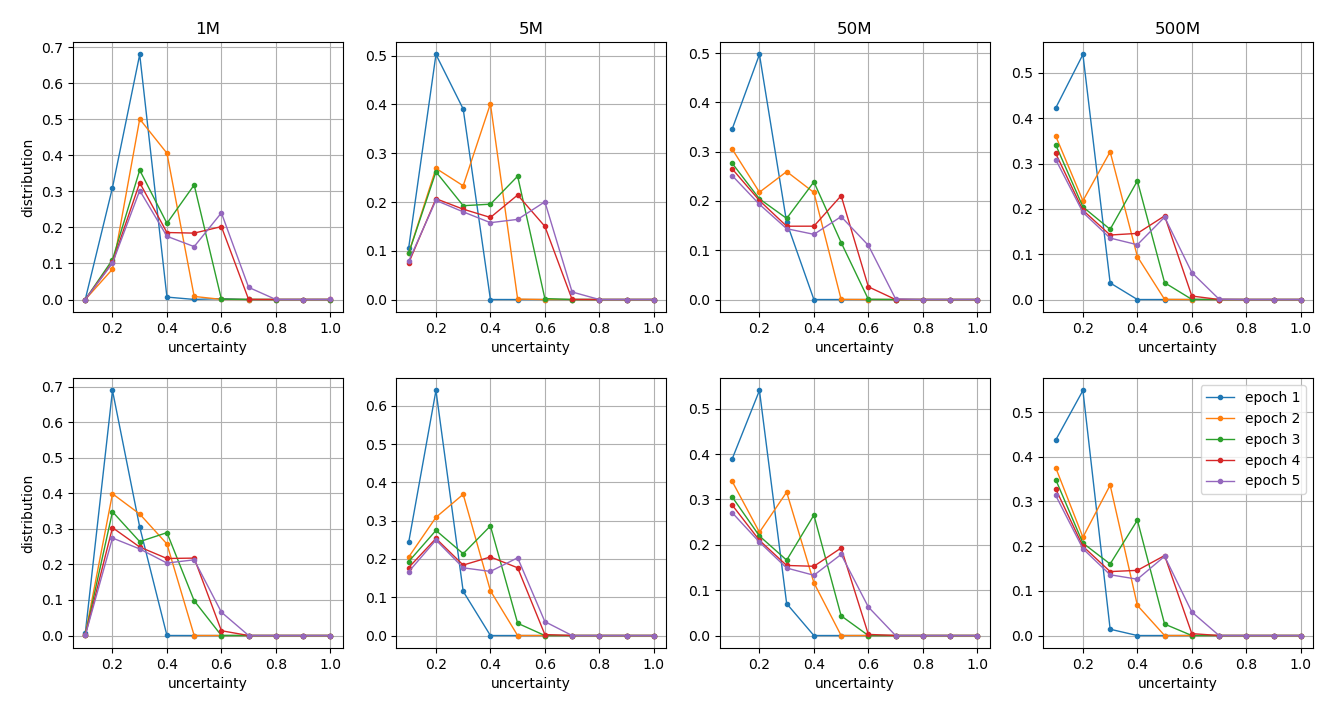}
\end{center}
   \caption{Uncertainty distribution of EDL (first row) and TEDL (second row), learnt on 1M, 5M and 50M training samples with $\lambda=0.1$. The first row is for EDL, while the second row is for TEDL.}
   \label{fig.exp_distribution}
\end{figure*}

\section{Experiments}
\label{sec_exp}

% This section will present our main experimental results, including a study on the sensitivity of the proposed TEDL method towards hyper-parameters. 

%--------------------------------------------------------------------------------------------------------------------

\subsection{Implementation Details}
\label{sec_sec_implementation}

All experiments throughout this paper are conducted on a binary classification task, with the goal to predict whether a <\textit{query}, \textit{ad}> pair is relevant or not. Both the training (1.4M) and validation samples (100K) are sampled from a large-scale commercial search engine, with human-provided relevance labels. In order to examine the impact of the size of training data, we further create a synthetic training set with soft labels, by sampling a large corpus and inference using an ensemble of BERT \cite{devlin2018bert} models fine-tuned on the human-labeled training set, similar to what we do in knowledge distillation \cite{li2019learning}. This allows us to experiment on a much larger scale, without breaking any assumptions in the EDL method. Without further clarification, we will henceforth refer to this synthetic training set as our training corpus, and experiments will be conducted on subsets sampled from this synthetic training set, with 1M, 5M, 50M and 500M samples, respectively. 

In addition, in this paper we will use TwinBERT \cite{lu2020twinbert, zhu2021textgnn} as our deep classification model, which uses two BERT encoders to encode \textit{query} and \textit{ad} respectively, and then calculates their relevance score by cosine similarity. We choose this model mainly for its simplicity and efficiency, and the conclusions of this paper should hold for other model architectures as well, since no particular assumptions for model architectures are made in the proposed TEDL method.

In terms of metrics, since we are working on binary classification task, we will use \textit{ROC AUC} to evaluate the prediction performance (in our experiments \textit{PR AUC} shows a very similar trend to \textit{ROC AUC}). Meanwhile, to measure the quality of uncertainty, we follow the approach in \cite{sensoy2018evidential} to split our validation data using different uncertainty thresholds first, and then evaluate \textit{ROC AUC} on each individual subset. For example, when threshold is $0.1$, \textit{ROC AUC} will be calculated only on validation samples with uncertainty lower than $0.1$. Therefore, if uncertainty is properly quantified, we should expect higher \textit{ROC AUC} on lower thresholds, since this is the subset that our model feels more confident with. This way, we can plot a curve over \textit{ROC AUC} v.s. uncertainty thresholds.

%--------------------------------------------------------------------------------------------------------------------

\subsection{Results and Analysis}
\label{sec_sec_results}

%--------------------------------------------------------------------------------------------------------------------

\subsubsection{Classification Performance evaluated by ROC AUC}
\label{sec_sec_sec_auc}

Figure \ref{fig.exp_auc} summarizes the per-epoch \textit{ROC AUC} of models learnt by cross-entropy loss, EDL method and the proposed TEDL method, with 1M, 5M and 50M training samples respectively. In all the three settings, we consistently observe that the \textit{ROC AUC} from EDL method is much lower than that from cross-entropy loss, while the proposed TEDL method is able to achieve comparable performance than cross-entropy loss, outperforming EDL significantly.

In addition, if we look into \textit{ROC AUC} measured on different epochs in Figure \ref{fig.exp_auc}, we can also see that TEDL is much more stable than EDL, especially when training corpus is relatively small.

%--------------------------------------------------------------------------------------------------------------------

\subsubsection{Quality of Uncertainty}
\label{sec_sec_sec_distribution}

As mentioned previously, we will measure the quality of the learnt uncertainty by plotting a curve over \textit{ROC AUC} v.s. uncertainty thresholds, as shown in Figure \ref{fig.exp_auc_u}, where the first row corresponds to EDL, while the second row is for TEDL. By comparing plots from different epochs, we can see that the quality of uncertainty learnt from both EDL and TEDL gets steadily improved over the training process, and the improving pattern for EDL and TEDL are very similar. However, this only happens when a relatively small $\lambda$ is used. Later in Section \ref{sec_sec_sensitivity} we will see that compared with EDL, TEDL is much more robust towards $\lambda$. We also plot the distribution of uncertainty in each training epoch, as shown in Figure \ref{fig.exp_distribution}, where TEDL also looks similar to EDL when $\lambda$ is relatively small, but later in Section \ref{sec_sec_sensitivity} we will see their difference when $\lambda$ gets larger.

%--------------------------------------------------------------------------------------------------------------------
\subsection{Sensitivity towards Hyper-parameters}
\label{sec_sec_sensitivity}

So far all the results we report are obtained under mild conditions with $\lambda=0.1$, however as we mentioned in Section \ref{sec_intro}, $\lambda$ and the number of training epochs may have dramatic impact on EDL, and hence it is necessary to examine how robust TEDL is towards these two hyper-parameters.

\begin{figure*}[h]
\begin{center}
   \includegraphics[width = 1.0\linewidth]{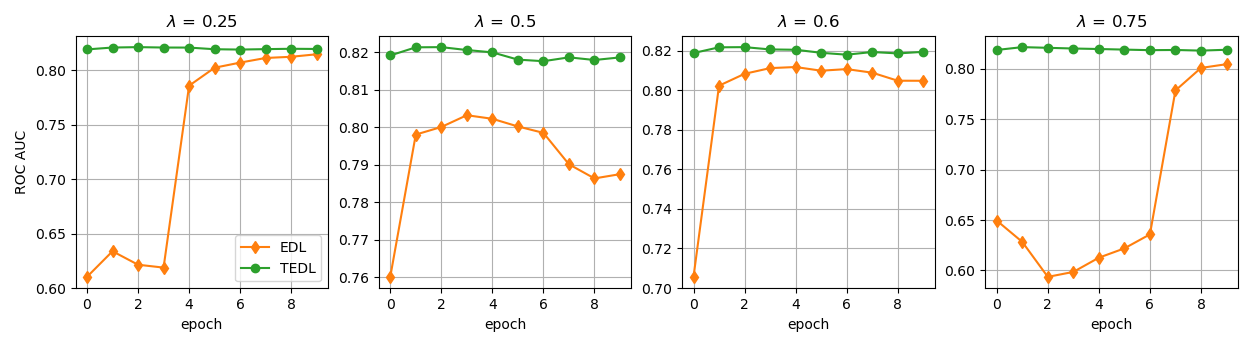}
\end{center}
   \caption{Comparison of \textit{ROC AUC} for EDL and TEDL, learnt on 1M training corpus with different $\lambda$. Compared to EDL, TEDL not only achieves higher \textit{ROC AUC}, but also shows improved robustness towards $\lambda$, especially when $\lambda=0.75$ where EDL method runs into the \textit{Dying ReLU} problem.}
   \label{fig.sensitivity_auc}
\end{figure*}

\begin{figure*}[!h]
\begin{center}
   \includegraphics[width = 0.97\linewidth]{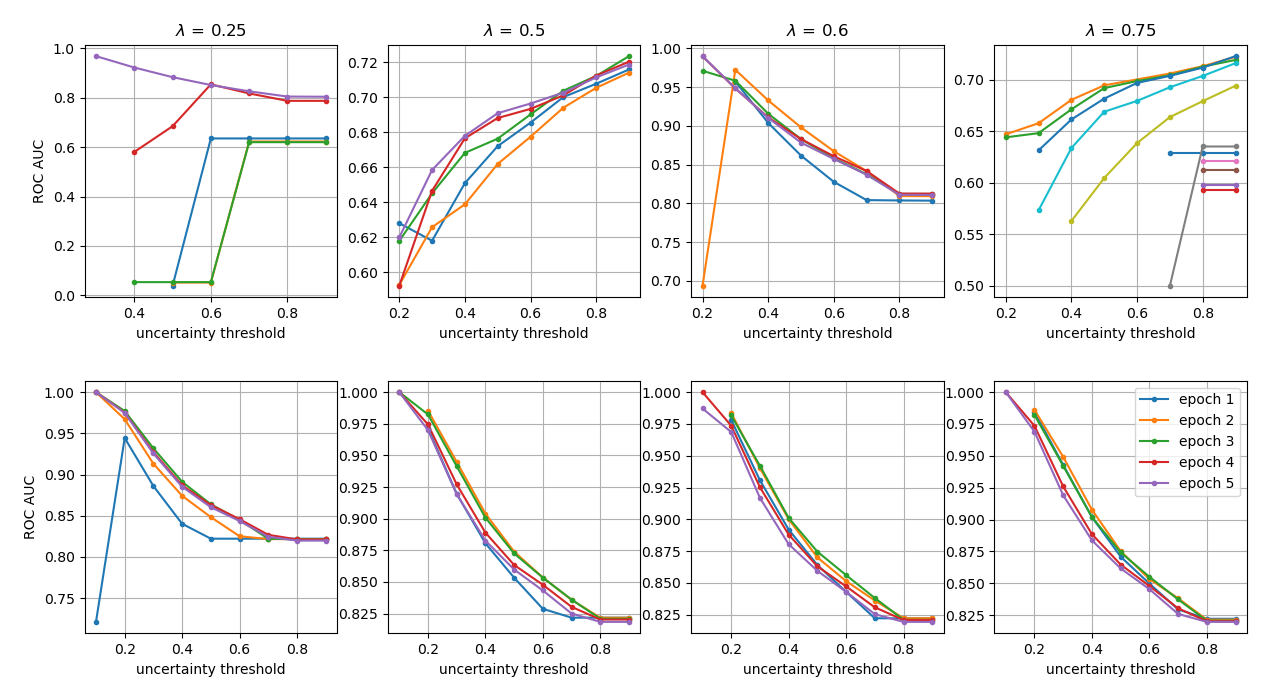}
\end{center}
   \caption{Comparison of \textit{ROC AUC} vs. uncertainty for EDL (first row) and TEDL (second row), learnt on 1M training corpus with different $\lambda$, where TEDL shows significantly better robustness.}
   \label{fig.sensitivity_auc_u}
\end{figure*}

\begin{figure*}[!h]
\begin{center}
   \includegraphics[width = 0.97\linewidth]{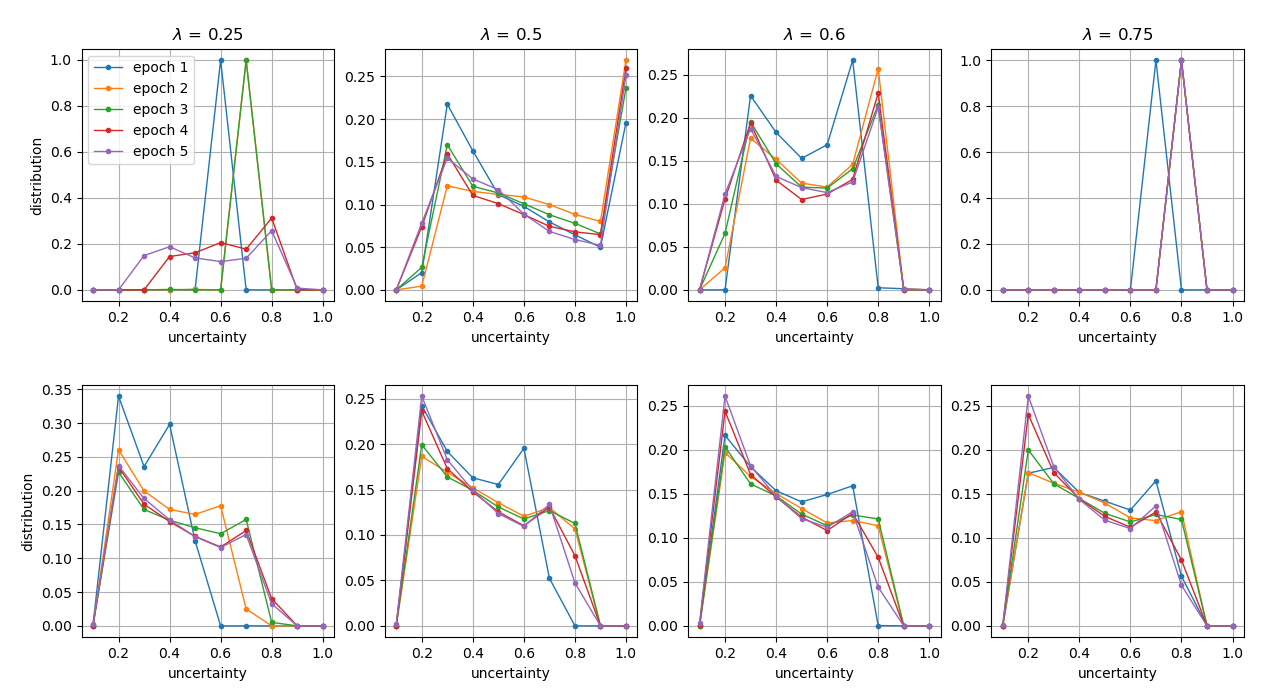}
\end{center}
   \caption{Comparison of uncertainty distribution for EDL (first row) and TEDL (second row), learnt on 1M training corpus with different $\lambda$, where TEDL shows significantly better robustness.}
   \label{fig.sensitivity_distribution}
\end{figure*}

%--------------------------------------------------------------------------------------------------------------------
\subsubsection{\textit{ROC AUC}}
\label{sec_sec_sec_sauc}

% \textbf{\textit{ROC AUC:}} 
Figure \ref{fig.sensitivity_auc} compares the \textit{ROC AUC} obtained by EDL and TEDL method, respectively, under different $\lambda$ values. Similar to Figure \ref{fig.exp_auc}, TEDL constantly outperforms EDL, and is more stable when more training epochs are used. In particular, when $\lambda=0.75$ we observe the \textit{Dying ReLU problem} in EDL, which inspires us to replace \textit{ReLU} by \textit{ELU} in TEDL.

%--------------------------------------------------------------------------------------------------------------------
\subsubsection{Quality of Uncertainty}
\label{sec_sec_sec_saucu}

Figure \ref{fig.sensitivity_auc_u} and Figure \ref{fig.sensitivity_distribution} compare the quality of uncertainty learnt by EDL and TEDL method, respectively, under different $\lambda$ values. Compared with Figure \ref{fig.exp_auc_u} and Figure \ref{fig.exp_distribution}, the uncertainty quality learnt from EDL degrades dramatically when larger $\lambda$ is used, as shown in the case where $\lambda=0.25$ and $\lambda=0.5$. By contrast, for TEDL, both its plots over \textit{ROC AUC} vs. uncertainty as well as its uncertainty distribution look very similar to what we observed for $\lambda=0.1$, demonstrating  significantly improved robustness towards $\lambda$. 

% \textbf{\textit{ROC AUC} vs. uncertainty: }

%--------------------------------------------------------------------------------------------------------------------
% \subsubsection{Uncertainty distribution}
% \label{sec_sec_sec_sdistribution}

% \textbf{Uncertainty distribution: }

%--------------------------------------------------------------------------------------------------------------------

\section{Conclusion}
\label{sec_conclusion}

In this paper, we propose TEDL, a two-stage learning approach to quantify uncertainty for deep classification models. TEDL contains two stages: the first stage learns from cross-entropy loss to obtain a good point estimate of the Dirichlet prior distribution, and then the second stage learns to quantify uncertainty via the reformulated EDL loss. We conduct extensive experiments using training corpus sampled from a real commercial search engine, which demonstrates that compared with EDL, the proposed TEDL not only achieves higher \textit{AUC}, but also shows improved robustness towards hyper-parameters. As future work, the uncertainty learnt by TEDL may be leveraged to developing active learning algorithms. 

%--------------------------------------------------------------------------------------------------------------------

\clearpage
% \newpage

\bibliographystyle{ACM-Reference-Format}
\bibliography{mybib}

% \appendix

\end{document}